\documentclass{article} % For LaTeX2e
\usepackage{iclr2019_conference,times}

\usepackage{amsmath,amsfonts,bm}

\def\eqref#1{equation~\ref{#1}}
\def\1{\bm{1}}

\DeclareMathAlphabet{\mathsfit}{\encodingdefault}{\sfdefault}{m}{sl}
\SetMathAlphabet{\mathsfit}{bold}{\encodingdefault}{\sfdefault}{bx}{n}

\usepackage{hyperref}
\usepackage{url}
\usepackage{graphicx}
\usepackage{booktabs}

\usepackage{multirow}
\title{Characterizing Audio Adversarial Examples Using Temporal Dependency}

\author{
	Zhuolin Yang \\
	Shanghai Jiao Tong University\\
	\And	
	Bo Li\\
	University of Illinois at Urbana--Champaign\\
		\and
\textbf{Pin-Yu Chen}\\
IBM Research\\
    \And
    Dawn Song \\
	UC, Berkeley\\
}

\newif\ifsubmit

\submitfalse

\ifsubmit
\newcommand{\bo}[1]{}
\newcommand{\dawn}[1]{}
\newcommand{\wh}[1]{}
\else
\newcommand{\bo}[1]{{\textcolor{blue}{[Bo: #1]}}}
\newcommand{\dawn}[1]{{\textcolor{purple}{[Dawn: #1]}}}
\newcommand{\wh}[1]{{\textcolor{orange}{[Warren: #1]}}}
\fi

\newcommand{\sk}{$S_k$}
\newcommand{\wholek}{$S_{\{whole, k\}}$}
\newcommand{\slk}{$S_{k^-}$}

\begin{document}

\maketitle

\begin{abstract}
Recent studies have highlighted adversarial examples as a ubiquitous threat to different neural network models and many downstream  applications. Nonetheless, as unique data properties have inspired distinct and powerful learning principles, this paper aims to explore their potentials towards mitigating adversarial inputs. In particular, our results reveal the importance of using the temporal dependency in audio data to gain discriminate power against adversarial examples. Tested on the automatic speech recognition (ASR) tasks and three recent audio adversarial attacks, we find that (i) input transformation developed from image adversarial defense provides limited robustness improvement and is subtle to advanced attacks; (ii) temporal dependency can be exploited to gain discriminative power against audio adversarial examples and is resistant to adaptive attacks considered in our experiments. Our results not only show promising means of improving the robustness of ASR systems, but also offer novel insights in exploiting domain-specific data properties to mitigate negative effects of adversarial examples.
\end{abstract}

\section{Introduction}
Deep Neural Networks (DNNs) have been widely adopted in a variety of machine learning applications~\citep{krizhevsky2012imagenet,hinton2012deep,levine2016end}.
However, recent work has demonstrated that DNNs are vulnerable to adversarial perturbations~\citep{szegedy2014intriguing,goodfellow2014explaining}.
An adversary can add negligible perturbations to inputs and generate adversarial examples to mislead DNNs, first found in image-based machine learning tasks \citep{goodfellow2014explaining,carlini2016towards,liu2016delving,chen2017ead,chen2017show,su2018robustness}.

Beyond images, given the wide application of DNN-based audio recognition systems, such as Google Home and Amazon Alexa, audio adversarial examples have also been studied recently~\citep{carlini2018audio,alzantot2018did,cisse2017houdini,Kreuk2018Fooling}. Comparing between image and audio learning tasks, although their state-of-the-art DNN architectures are quite different (i.e., convolutional v.s. recurrent neural networks), the attacking methodology towards generating adversarial examples is fundamentally unanimous - finding adversarial perturbations through the lens of maximizing the training loss or optimizing some designed attack objectives. For example, the same attack loss function proposed in \citep{cisse2017houdini} is used to generate adversarial examples in both visual and speech recognition models.
Nonetheless, different types of data usually possess unique or domain-specific properties that can potentially be used to gain discriminative power against adversarial inputs. In particular, the temporal dependency in audio data is an innate characteristic
that has already been widely adopted in the machine learning models. However, in addition to improving learning performance on natural audio examples, it is still an open question on whether or not the temporal dependency can be exploited to help mitigate negative effects of adversarial examples.

The focus of this paper has two folds. First, we investigate the robustness of automatic speech recognition (ASR) models under  \textit{input transformation}, a commonly used technique in the image domain to mitigate adversarial inputs.
Our experimental results show that four implemented transformation techniques on audio inputs, including waveform quantization, temporal smoothing, down-sampling and autoencoder reformation, provide limited robustness improvement against the recent attack method proposed in \citep{athalye2018obfuscated}, which aims to circumvent the gradient obfuscation issue incurred by input transformations.
Second, we demonstrate that temporal dependency can be used to gain discriminative power against adversarial examples in ASR.
We perform the proposed temporal dependency method on both the LIBRIS~\citep{graetz1986application} and Mozilla Common Voice datasets against three state-of-the-art attack methods~\citep{carlini2018audio,alzantot2018did,yuan2018commandersong} considered in our experiments and show that such an approach achieves promising identification of non-adaptive and adaptive attacks. %\bo{do we want to add some results for commandsong and did you hear that?}
Moreover, we also verify that the proposed method can resist strong proposed adaptive attacks in which the defense implementations are known to an attacker. Finally, we note that although this paper focuses on the case of audio adversarial examples, the methodology of leveraging unique data properties to improve model robustness could be naturally extended to different domains. The promising results also shed new lights in designing adversarial defenses against attacks on various types of data.

\textbf{Related work }
% Background on Audio Adversarial Examples
     An adversarial example for a neural network is an input $x_{adv}$ that is similar to a natural input $x$ but will yield different output after passing through the neural network. Currently, there are two different types of attacks for generating audio adversarial examples: the Speech-to-Label attack and the Speech-to-Text attack.
    The Speech-to-Label attack aims to find an adversarial example $x_{adv}$ close to the original audio $x$ but yields a different (wrong) label. To do so, Alzantot et al.  proposed a genetic algorithm \citep{alzantot2018did}, and Cisse et al.  proposed a probabilistic loss function \citep{cisse2017houdini}.
    The Speech-to-Text attack requires the transcribed output of the adversarial audio to be the same as the desired output, which has been made possible by Carlini and Wagner \citep{carlini2018audio} using
    optimization-based techniques operated on the raw waveforms.  Iter et al. leveraged extracted audio features called Mel Frequency Cepstral Coefficients (MFCCs)  \citep{iter2017generating}.
    Yuan et al. demonstrated practical ``wav-to-API'' audio adversarial attacks \citep{yuan2018commandersong}. Another line of research focuses on adversarial training or data augmentation to improve model robustness \citep{serdyuk2016invariant,michelsanti2017conditional,sriram2017robust,sundata18}, which is beyond our scope. Our proposed approach focuses on gaining the discriminative power against adversarial examples through embedded temporal dependency, which is compatible with any ASR model and does not require adversarial training or data augmentation.
% \bo{we can decide if we wanna mention input pre-processing based defense here for images based on our page limit?}

\section{Do Lessons from Image Adversarial Examples Transfer to Audio Domain?}

Although in recent years both image and audio learning tasks have witnessed significant breakthroughs accomplished by advanced neural networks, these two types of data have unique properties that lead to distinct learning principles.
In images, the pixels entail spatial correlations corresponding to hierarchical object associations and color descriptions, which are leveraged by the convolutional neural networks (CNNs) for feature extraction. In audios, the waveforms possess apparent temporal dependency, which is widely adopted by the recurrent neural networks (RNNs).
For the segmentation task in image domain, spatial consistency has played an important role in improving model robustness~\citep{lowe1999object}. However, it remains unknown whether temporal dependency can have a similar effect of improving model robustness against audio adversarial examples.
In this paper, we aim to address the following fundamental questions: (a) \textit{do lessons learned from image adversarial examples transfer to the audio domain?}; and (b) \textit{can temporal dependency be used to discriminate audio adversarial examples?}
Moreover, studying the discriminative power of temporal dependency in audios not only highlights the importance of using unique data properties towards building robust machine learning models, but also aids in devising principles for investigating more complex data such as videos (spatial + temporal properties) or multimodal cases (e.g., images + texts).

Here we summarize two primary findings concluded from our experimental results in Section \ref{sec_experiment}.

\textbf{Audio input transformation is not effective against adversarial attacks} Input transformation is a widely adopted defense technique in the image domain, owing to its low operation cost and easy integration with the existing network architecture~\citep{luo2015foveation,wang2016using,dziugaite2016study}. Generally speaking, input transformation aims to perform certain feature transformation on the raw image in order to disrupt the adversarial perturbations before passing it to a neural network. Popular approaches include bit quantization, image filtering, image reprocessing, and  autoencoder reformation \citep{xu2017feature,guo2017countering,meng2017magnet}. However, many existing methods are shown to be bypassed by subsequent or adaptive adversarial attacks \citep{carlini2017adversarial,he2017adversarial,carlini2017magnet,lu2018limitation}. Moreover, Athalye et al. \citep{athalye2018obfuscated} has pointed out that input transformation may cause \textit{obfuscated gradients} when generating adversarial examples and thus gives a false sense of robustness. They also demonstrated that in many cases this gradient obfuscation issue can be circumvented, making input transformation still vulnerable to adversarial examples. Similarly, in our experiments we find that audio input transformations based on waveform quantization, temporal filtering, signal down sampling or autoencoder reformation suffers from similar weakness: the tested model with input transformation becomes fragile to adversarial examples when one adopts the attack considering gradient obfuscation as in \citep{athalye2018obfuscated}.

\textbf{Temporal dependency possesses strong discriminative power against adversarial examples in automatic speech recognition} Instead of input transformation, in this paper we propose to exploit the inherent temporal dependency in audio data to discriminate adversarial examples. Tested on the automatic speech recognition (ASR) tasks, we find that the proposed methodology can effectively detect audio adversarial examples while minimally affecting the recognition performance on normal examples. In addition, experimental results show that a considered adaptive adversarial attack, even when knowing every detail of the deployed temporal dependency method, cannot generate adversarial examples that bypass the proposed temporal dependency based approach.

Combining these two primary findings, we conclude that the weakness of defense techniques identified in the image case is very likely to be transferred to the audio domain. On the
other hand, exploiting unique data properties to develop defense methods, such as using temporal dependency in ASR, can lead to promising defense approaches that can resist adaptive adversarial attacks.
\section{Input Transformation and Temporal Dependency in Audio Data}
In this section, we will introduce the effect of basic input transformations on audio adversarial examples, and analyze temporal dependency in audio data. We will also show that such temporal dependency can be potentially leveraged to discriminate audio adversarial examples.

\subsection{ Audio Adversarial Examples Under Simple Input Transformations}

    Inspired by image input transformation methods and as a first attempt, we applied some primitive signal processing transformations to audio inputs. These transformations are useful, easy to implement, fast to operate and have delivered several interesting findings.

    \textbf{\textbf{Quantization:}} By rounding the amplitude of audio sampled data into the nearest integer multiple of $q$,  the adversarial perturbation could be disrupted since its amplitude is usually small in the input space. We choose $q = 128, 256, 512, 1024$ as our parameters.

    \textbf{\textbf{Local smoothing:}} We use a sliding window of a fixed length for local smoothing to reduce the adversarial perturbation. For an audio sample $x_i$, we consider the $K-1$ samples before and after it, denoted by $[x_{i-K+1},\dots,x_i,\dots,x_{i+K-1}]$, as a local reference sequence and replace $x_i$ by the smoothed value (average, median, etc) of its reference sequence.

    \textbf{\textbf{Down sampling:}} Based on sampling theory, it is possible to down-sample a band-limited audio file without sacrificing the quality of the recovered signal while mitigating the adversarial perturbations in the reconstruction phase. In our experiments, we down-sample the original 16kHz audio data to 8kHz and then perform signal recovery.

    \textbf{\textbf{Autoencoder:}} In adversarial image defending field, the MagNet defensive method \citep{meng2017magnet} is an effective way to remove adversarial noises: Implement an autoencoder to project the adversarial input distribution space into the benign distribution. In our experiments, we implement a sequence-to-sequence autoencoder, and the whole audio will be cut into frame-level pieces passing through the autoencoder and concatenate them in the final stage, while using the whole audio passing the autoencoder directly is proved to be ineffective and hard to utilize the underlying information.
    % The diagram of Magnet transformation are shown in \ref{tab:Fig2}.

%\begin{figure}[htbp]
    %\centering
    %\includegraphics[height=6cm]{nips18_audio_temporalConsistency/magnet.pdf}
    %\caption{Magnet based audio transformation diagram}
    %\label{tab:Fig2}
%\end{figure}

\subsection{Temporal Dependency Based Method (TD)}
\begin{figure}[t]
   \vspace{-6mm}
    \centering
    \includegraphics[width=1\textwidth]{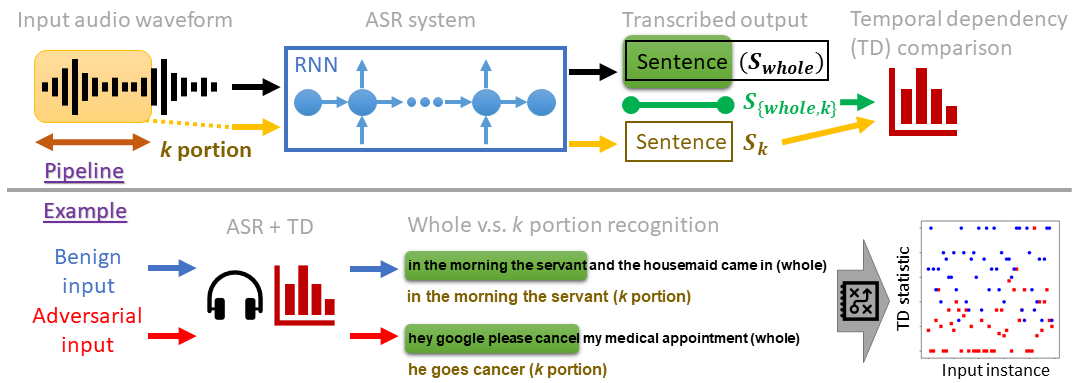}
    \caption{Pipeline and example of the proposed temporal dependency (TD) based method for discriminating audio adversarial examples.}
    \label{tab:Fig1}
    \vspace{-6mm}
\end{figure}

Due to the fact that audio sequence has explicit temporal dependency (e.g., correlations in consecutive waveform segments), here we aim to explore if such temporal dependency will be affected by adversarial perturbations.
The pipeline of the temporal dependency based method is shown in Figure~\ref{tab:Fig1}. Given an audio sequence, we propose to select the first $k$ portion of it (i.e., the prefix of length $k$) as input for ASR to obtain transcribed results as \sk. We will also insert the whole sequence into ASR and select the prefix of length $k$ of the transcribed result as \wholek, which has the same length as \sk. We will then compare the consistency between \sk and \wholek in terms of temporal dependency distance. Here we adopt the word error rate (WER) as the distance metric~\citep{Levenshtein1966Binary}.
For normal/benign audio instance, \sk~and \wholek~should be similar since the ASR model is consistent for different sections of a given sequence due to its temporal dependency.
However, for audio adversarial examples, since the added perturbation aims to alter the ASR ouput toward the targeted transcription, it may fail to preserve the temporal information of the original sequence. Therefore, due to the loss of temporal dependency, \sk~and \wholek~in this case will not be able to produce consistent results.
Based on such hypothesis, we leverage the prefix of length $k$ of the transcribed results and the transcribed  $k$ portion to potentially recognize adversarial inputs.

\section{Experimental Results}
\label{sec_experiment}
The presentation flows of the experimental results are summarized as follows. We will first introduce the datasets, target learning models, attack methods, and evaluation metrics for different defense/detection methods that we focus on.
We then discuss the defense/detection effectiveness for different methods against each attack respectively. Finally, we evaluate strong adaptive attacks against these defense/detection methods.
We show that due to different data properties, the autoencoder based defense cannot effectively recover the ground truth for adversarial audios and may also have negative effects on benign instances as well. Input transformation is less effective in defending adversarial audio than images. In addition, even when some input transformation is effective for recovering some adversarial audio data, we find that it is easy to perform adaptive attacks against them.
The proposed TD method can effectively detect adversarial audios generated by different attacks targeting on various learning tasks (classification and speech-to-text translation).
In particular, we propose different types of strong adaptive attacks against the TD detection method. We show that these strong adaptive attacks are not able to generate effective adversarial audio against TD and we provide some case studies to further understand the performance of TD.

%\textcolor{blue}{We need to introduce Datasets, Model, and Metrics (WER CER ratio) upfront.}
\subsection{Experimental setup}

In our experiments, we measure the effectiveness on several adversarial audio generation methods. For audio classification attack, we used Speech Commands dataset. For speech-to-text attack, we benchmark each method on both LibriSpeech and Mozilla Common Voice dataset. In particular, for the Commander Song attack \citep{yuan2018commandersong}, we measure on the generated adversarial audios given by the authors.

\textbf{Dataset}

\textit{LibriSpeech dataset}: LibriSpeech~\citep{Panayotov2015Librispeech} is a corpus of approximately 1000 hours of 16Khz English speech derived from audiobooks from the LibriVox project. We used samples from its test-clean dataset in their website and the average duration is 4.294s. We generated adversarial examples using the attack method in~\citep{carlini2018audio}.

\textit{Mozilla Common Voice dataset}: Common Voice is a large audio dataset provided by Mozilla. This dataset is public and contains samples from  human  speaking audio files. We used the 16Khz-sampled data released in \citep{carlini2018audio}, whose average duration is 3.998s. The first 100 samples from its test dataset is used to mount attacks, which is the same attack experimental setup as in~\citep{carlini2018audio}.

Speech Commands dataset: Speech Commands dataset~\citep{Warden2018Speech} is a audio dataset contains 65000 audio files. Each audio is just a single command lasting for one second. Commands are "yes", "no", "up", "down", "left", "right", "on", "off", "stop", and "go".

\textbf{Model and learning tasks}
%In our proposed method, it is flexible to choose different models to compute and compare the temporal consistency. Without loss of generality, in our evaluation, we leverage the commonly used model for each task.
For speech-to-text task, we use DeepSpeech speech-to-text transcription network, which is a biRNN based model with beam search to decode text. For audio classification task, we use a convolutional speech commands classification model. For the Command Song attack, we evaluate the performance on Kaldi speech recognition platform.
%we applied our method directly on adversarial audio and evaluate them on Kaldi speech recognition platform.

\textbf{Attack Methods}
%\bo{Zhuolin, can you cite and add here.}

\emph{Genetic algorithm based attack against audio classification (GA)}: For audio classification task, we consider the state-of-the-art attack proposed in~\citep{alzantot2018did}.
Here an audio classification model is attacked and the audio classes include ``yes, no, up, down, etc.". They aimed to attack such a network to misclassify an adversarial instance based on either
targeted or untargeted attack.

\emph{Commander Song attack against speech-to-text translation (Commander)}: Commander Song  \citep{yuan2018commandersong} is a speech-to-text targeted attack which can attack an audio extracted from a popular song. The adversarial audio can even be played over the air with its adversarial characteristics. Since the Commander Song codes are not available, we  measure the effectiveness of the generated adversarial audios given by the authors.

\emph{Optimization based attack against speech-to-text translation (Opt)}: We consider the targeted speech-to-text attack proposed by~\citep{carlini2018audio}, which uses CTC-loss in a speech recognition system as an objective function and solves the task of adversarial attack as an optimization problem.

\textbf{Evaluation Metrics}
For \emph{defense method} such as input transformation, since it aims to recover the ground truth (original instances) from adversarial instances, we use the word error rate (WER) and character error rate (CER)~\citep{Levenshtein1966Binary} as evaluation metrics to measure the recovery efficiency.
WER and CER are commonly used metrics to measure the error between recovered text and the ground truth in word level or character level. Generally speaking, the error rate (ER) is defined by $ER = \frac{S+D+I}{N}$, where $S, D, I$ is the number of substitutions, deletions and insertions calculated by dynamic string alignment, and $N$ is the total number of word / character in the ground truth text. %\bo{we need to cover the three metrics in table 1}

To fairly evaluate the effectiveness of these transformations against speech-to-text attack, we also report the ratio of translation distance between instance and corresponding ground truth before and after transformation.
For instance, as a controlled experiment, given a audio instance $x$ (adversarial instance is denoted as $x_{adv}$), its corresponding ground truth $y$, and the ASR function $g(\cdot)$, we calculate the effectiveness ratio for benign instances as $R_{benign}=\frac{D(g(T(x)), y)}{D(g(x), y)}$,
where $T(\cdot)$ denotes the result of transformation and $D(\cdot,\cdot)$ characterizes the distance function (WER and CER in our case).
For adversarial audio, we calculate the similar effectiveness ratio as $R_{adv}=\frac{D(g(T(x_{adv})), y)}{D(g(x_{adv}), y)}$.
%\bo{check here whether we mention every metric in the table}

For \emph{detection method}, the standard evaluation metric is the area under curve (AUC) score, aiming to evaluate the detection efficiency.
The proposed TD method is the first data-specific metric to detect adversarial audio, which focuses on how many adversarial instances are captured (true positive) without affecting benign instances (false positive). Therefore, we follow the standard criteria and report AUC for TD. For the proposed TD method, we compare the temporal dependency based on WER, CER, as well as the longest common prefix (LCP).
LCP is a commonly used metric to evaluate the similarity between two strings. Given strings $b_1$ and $b_2$, the corresponding LCP is defined as $max_{b_1[:k] = b_2[:k]}~k$, where $[:k]$ represents the prefix of length $k$ of a translated sentence.

\subsection{Evaluation of Defense methods against adversarial audio}
In this section we measured our defense method of audoencoder based defense and input transformation defense for classification attack (GA) and speech-to-text attack (Commander and Opt). We summarize our work in Table~\ref{tab:finaltable1} and list some basic results. For Commander, due to  unreleased training data, we are not able to train an autoencoder. For GA and Opt we have sufficient data to train autoencoder. %\textcolor{blue}{And for classification attack, because of the speechcommands only contains one word transcribe text and it's easy to defense under some simple input transformation method, we didn't actually evaluate the detection performance of classification attack. (PY: should be removed)}
\begin{table}[]
\caption{List of adversarial audio based attacks and corresponding evaluation results for defense and detection methods}
\vspace{-4mm}
\label{tab:finaltable1}
\begin{center}
\begin{small}
\resizebox{1.00\textwidth}{19mm}{
\begin{tabular}{|c|c|c|c|c|c|c|}
\toprule
\multicolumn{3}{|c|}{Learning tasks}                                                                                                                                & Classification                                                                & \multicolumn{3}{c|}{Speech-to-Text}                                                                                                                                   \\ \hline
\multicolumn{3}{|c|}{Attack methods}                                                                                                                                & Genetic Algorithm (GA)                                                             & CommanderSong (Commander)                                                            & \multicolumn{2}{c|}{Opt. attack (Opt)}                                                   \\ \hline

\multicolumn{3}{|c|}{Datasets}                                                                                                                                      & SpeechCommand                                                                    & Some popular songs                                                        & LibriSpeech                                     & CommonVoice                                    \\ \hline
\multicolumn{3}{|c|}{Evaluation metrics of defense}                                                                                                              & \begin{tabular}[c]{@{}c@{}}Average attack \\ success rate\end{tabular} & \begin{tabular}[c]{@{}c@{}}Target command\\ recognition rate\end{tabular} & \multicolumn{2}{c|}{\begin{tabular}[c]{@{}c@{}}Effectiveness ratio\\ $\text{R}_{benign}$ / $\text{R}_{adv}$\end{tabular}} \\ \hline
\multirow{5}{*}{\begin{tabular}[c]{@{}c@{}}Defense\\ Method\end{tabular}} & \multicolumn{2}{c|}{Without defense}                                                 & 84\%                                                                   & 100\%                                                                     & 1.0 / 1.0                                            & 1.0 / 1.0                                        \\ \cline{2-7}
                                                                          & \multirow{3}{*}{\begin{tabular}[c]{@{}c@{}}Input\\ trans.\end{tabular}} & Down Samp. & 3.2\%                                                                  & 8\%                                                                       & 3.87 / 0.40                                          & 1.13 / 0.73                                      \\ \cline{3-7}
                                                                          &                                                                         & Quan-256   & 2.1\%                                                                  & 4\%                                                                       & 1.13 / 0.28                                          & 1.56 / 0.74                                      \\ \cline{3-7}
                                                                          &                                                                         & Median-4   & 2.7\%                                                                  & 4\%                                                                       & 1.18 / 0.34                                          & 0.98 / 0.77                                      \\ \cline{2-7}
                                                                          & \multicolumn{2}{c|}{Autoencoder}                                                     & 8.2\%                                                                  & -                                                                         & 9.84 / 0.97                                          & 2.09 / 0.80                                      \\ \hline

\multicolumn{3}{|c|}{Evaluation metrics for detection}                                                                                                              & -                                                                              & Detection rate                                                            & \multicolumn{2}{c|}{AUC score}                                                                   \\ \hline
\multicolumn{3}{|c|}{Detection results of TD Method}                                                                                                                & -                                                                              & \textbf{1.00}                                                                     & \textbf{0.930}                                           & \textbf{0.936}                                          \\ \bottomrule

\end{tabular}}
\end{small}
\end{center}
\vspace{-6mm}
\end{table}
% \usepackage{multirow}
% \usepackage{multirow}

\textbf{Input Transformation as Defense}
%Currently, there are two types of audio attacks: attacking audio classification and attacking speech-to-text tasks. We will first analyze the effect of various input transformations on different attacks.

% We can guess that because the cutting into frame-level audio process break the temporal dependency and cause the magnet ways failed.

Here we perform the primitive input transformation for audio classification targeted attacks and evaluate the corresponding effects. Due to the space limitation, we defer the results of untargeted attacks to the supplemental materials.

\begin{figure}[htbp]
    \centering
    \vspace{-5mm}
    \begin{minipage}[h]{0.45\textwidth}
    \centering
        \includegraphics[height=4.5cm]{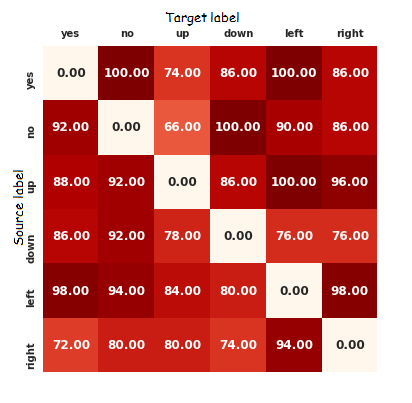}
        \vspace{-5mm}
        \caption{Attack success rates (\%) }
        \label{tab:smallPic1}
    \end{minipage}
    \begin{minipage}[h]{0.5\textwidth}
        \centering
        \includegraphics[height=4.5cm]{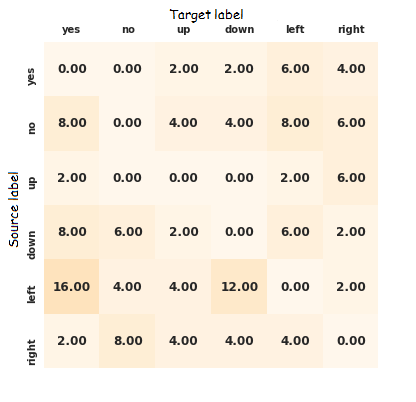}
        \vspace{-5mm}
        \caption{Attack success (\%) after transformation}
        \label{tab:smallPic2}
    \end{minipage}
    \vspace{-4mm}
\end{figure}

\emph{\textbf{GA}} We first evaluate our input transformation against the audio classification attack (GA) in \citep{alzantot2018did}. We implemented their attack with 500 iterations and limit the magnitude of adversarial perturbation within 5 (smaller than the quantization we used in transformation) and generated 50 adversarial examples per attack task (more targets are shown in supplementary material). The attack success rate is $84\%$ on average. For the ease of illustration, we use Quantization-256 as our input transformation.  As observed in Figures~\ref{tab:smallPic1} and \ref{tab:smallPic2}, the attack success rates decreased to only $2.1\%$, and $63.8\%$ of the adversarial instances have been converted back to their original (true) label. We also measure the possible effects on original audio due to our transformation methods: the original audio classification accuracy without our transformation is $89.2\%$, and the rate slightly decreased to $89.0\%$ after our transformation, which means the effects of input transformation on benign instances are negligible.
In addition, it also shows that for classification tasks, such input transformation is more effective in mitigating negative effects of adversarial perturbation. This potential reason could be that classification tasks do not rely on audio temporal dependency but focuses on local features, while speech-to-text task will be harder to defend based on the tested input transformations.
% \bo{do we want to put more analysis here?}

\emph{\textbf{Commander}} We also evaluate our input transformation method against the Commander Song attack~\citep{yuan2018commandersong}, which implemented an Air-to-API adversarial attack. In the paper, the authors reported  $91\%$ attack detection rate using their defense method. We measured our Quan-256 input transformation on 25 adversarial examples obtained via personal communications. Based on the same detection evaluation metric in \citep{yuan2018commandersong}\footnote{The authors set the detection threshold to be 0 and we used the same setting here.}, Quan-256 attains 100\% detection rate for characterizing all the adversarial examples.

\emph{\textbf{Opt} }
%In addition to autoencoder, here we study the effects of other general primitive transformations on benign and adversarial audio.
Here we consider the state-of-the-art audio attack proposed in \citep{carlini2018audio}. We separately choose 50 audio files from two audio datasets (Common Voice, LIBRIS) and generate attacks based on the CTC-loss. We evaluate several primitive signal processing methods as input transformation under WER and CER metrics in Table~\ref{tab:commondeflanguage} and \ref{tab:librideflanguage}. We then also evaluate the WER and CER based effectiveness ratio we mentioned before to Quanify the effectiveness of transformation. $R_{benign}$ are shown in the brackets for the first two columns in Table~\ref{tab:commondeflanguage} and \ref{tab:librideflanguage}, while $R_{adv}$ is shown
in the brackets of last two columns within those tables.
We compute our results using both ground truth and adversarial target ``This is an adversarial example" as references.
% metrics and also report the ratio between the transcribing error rate referenced by ground truth and the adversarial target (shown in brackets).  In these table, we showed the WER and CER scores using ground truth sentence as reference and we also list the ratio between the WER / CER scores of transformed audio (benign or adversarial) and audios without transformation and this ratio can showed a lot of thing. More specifically:

% For evaluation of input transformation on benign audio:
%     \[ratio = \frac{ref(x, \textup{ground truth})}{ref(\textup{benign}, \textup{ground truth})}\]

% and for input transformation on adversarial audio:
%     \[ratio = \frac{ref(x, \textup{ground truth})}{ref(\textup{adversarial}, \textup{ground truth})}\]
Here small $R_{benign}$ which is close to 1 indicates that transformation has little effect on benign instances, small $R_{adv}$ represents transformation is effective recovering adversarial audio back to benign. From Tables~\ref{tab:commondeflanguage} and \ref{tab:librideflanguage} we showed that most of input transformations (e.g., Median-4, Downsampling and Quan-256) effectively reduce the adversarial perturbation without affecting the original audio too much.

Although these input transformations show certain effectiveness in defending against adversarial audios, we find that it is still possible to generate adversarial audios by adaptive attack in Section~\ref{sec:adaptive}.
%However, as we can see that Autoencoder

\textbf{Autoencoder as Defense}

Towards defending against (non-adaptive) adversarial images, MagNet~\citep{meng2017magnet} has achieved promising performance by using an antoencoder to mitigate adversarial perturbation. Inspired by it, here we apply a similar autoencoder structure for audio and test if such input transformation can be applied to defending against adversarial audio.
% We aim to similarly build a network to learn the underlying feature of adversarial audio and turn these adversarial audio into benign audio.
% Corresponding to adversarial image defending field, Magnet is a effective way defending adversarial image by using antoencoder to erase adversarial noise.
We apply a MagNet-like method for feature-extracted audio spectrum map: we build an encoder to compress the information of origin audio features into latent vector $z$, then use $z$ for reconstruction  by passing through another decoder network under frame level and combine them to obtain the transformed audio~\citep{Hsu2017Unsupervised}. Here we analyzed the performance of Autoencoder transformation in both \emph{GA} and \emph{Opt} attack. We find that MagNet which gained great effectiveness on defending adversarial images in the oblivious attack setting \citep{carlini2017magnet,lu2018limitation}, has limited effect on audio defense.

\emph{\textbf{GA}}
We presented our results in Table~\ref{tab:finaltable1} that against classification attack, Autoencoder did not perform well by only reducing attack success rate to $8.2\%$ defeat by other input transformation methods. Since you can reduce attack success rate to $10\%$ by just destroying the origin audio data and altering to random guess, it's hard to say that Autoencoder method has good performance.

\emph{\textbf{Opt}}
We report that the autoencoder works not very well for transforming benign instances (57.6 WER in Common Voice compared to 27.5 WER without transformation, 30.0 WER in LIBRIS compared to 12.4 WER without transformation), also fails to recover adversarial audio (76.5 WER in Common Voice and 99.4 WER in LIBRIS). This shows that the non-adaptive additive adversarial perturbation can bypass the MagNet-like autoencoder on
audio, which implies different robustness implications of image and audio data.

\subsection{Evaluation of TD detection method against Adversarial audio}
In this section, we will evaluate the proposed TD detection method on different attacks. We will first report the AUC for detecting different attacks with TD to demonstrate the effectiveness, and we will provide some additional analysis and examples to help better understand TD.
We only evaluate our TD method on speech-to-text attacks (Commander and Opt) because the audio in the Speech Commands dataset for classification attack is just a single command lasting for one second and thus its temporal dependency is not obvious.

\emph{\textbf{Commander}}
In Commander Song attack,
we directly examine whether the generated adversarial audio is consistent with its prefix of length $k$ or not. We report that by using TD method with $k = \frac{1}{2}$, all the generated adversarial samples showed inconsistency and thus were successfully detected.

\emph{\textbf{Opt}}
Here we show the empirical performance of distinguishing adversarial audios by leveraging the temporal dependency of audio data.
% In the experiments, we use these three metrics: word error rate (WER), character error rate (CER)~\citep{Levenshtein1966Binary}, longest common prefix (LCP)~\citep{Edelsbrunner1986Optimal} to measure the inconsistency between \sk and \wholek.
In the experiments, we use these three metrics, WER, CER and LCP, to measure the inconsistency between \sk~and \wholek.
As a baseline, we also directly train a one layer LSTM with 64 hidden feature dimensions based on the collected adversarial and benign audio instances for classification.
Some examples of translated results for benign and adversarial audios are shown in Table~\ref{tab:examples}. Here we consider three types of adversarial targets: short -- \emph{hey google}; medium -- \emph{this is an adversarial example}; and long -- \emph{hey google please cancel my medical appointment}.
We report the AUC score for these detection results for $k=1/2$ in Table~\ref{tab:AUC}.
% our method's performance by using AUC scores. More specifically, Let $x$ be the input audio and $x^{\ast}$ be the first $k$-portion of $x$. Let $f(\cdot)$ denote the speech recognizing and transcribing process. To measure the similarity between $f(x)$ and $f(x^{\ast})$ (they usually have different lengths), we only compare the prefix $f^{\ast}(x)$ of $f(x)$ that has the same length as $f(x^{\ast})$ and ignore the rest. Thus, we applied our metrics (WER, CER, LCP) to measure the distance between $f^{\ast}(x)$ and $f(x^{\ast})$. To make the results comparable, We also add a trained baseline LSTM which has 64 hidden layer features as our baseline model.
% Results and some examples of $k = 1 / 2$ are shown in \ref{tab:AUC} and
% we can show that by using WER score as metric, it achieved AUC score 0.936 on Common Voice and 0.930 on LIBRIS. We also iterated different ratio $k$ based on three metric and results are shown in \ref{tab:parameter k}. We can find that when $k$ moves from $1 / 2$ to 1(full comparison), the AUC climbs up and then falls down (which is reasonable). When $k = 4 / 5$, it reaches the culmination with AUC scores $0.969$ by using CER metric, which is a surprisingly high result. The scatter map of benign audio and adversarial audio's detection value are shown on Fig. We can find that these results suggest an easy-implemented but effective method for characterizing adversarial audio attacks.

\begin{table}[thb]
\begin{small}
\vspace{-6mm}
        \caption{Examples of the temporal dependency based detection method}
        \label{tab:examples}
        \begin{center}
        \begin{small}
        \begin{tabular}{ll}
        \hline
        Type & Transcribed results \\
        \hline
        Original & then good bye said the rats and they went home \\
        the first half of Original & then good bye said the {\color{red}raps}\\ \\

        Adversarial (short) & hey google \\
        First half of Adversarial & {\color{red}he is}\\
        Adversarial (medium) & this is an adversarial example \\
        First half of Adversarial & {\color{red}thes on adequate}\\
        Adversarial (long) & hey google please cancel my medical appointment \\
        First half of Adversarial & {\color{red}he goes cancer}\\
        \hline
        \end{tabular}
        \end{small}
        \end{center}
        \end{small}
        \vspace{-2mm}
\end{table}

\begin{table}[t]
\begin{small}
\vspace{-4mm}
        \caption{AUC results of the proposed temporal dependency method}
        \label{tab:AUC}
        \begin{center}
        \begin{small}
        \begin{tabular}{ccccc}
        \hline
        Dataset & LSTM & TD (WER) & TD (CER) & TD (LCP ratio) \\
        \hline
        Common Voice & 0.712 & \textbf{0.936} & 0.916 & 0.859 \\
        LIBRIS & 0.645 & 0.930 & \textbf{0.933} & 0.806 \\
        \hline
        \end{tabular}
        \end{small}
        \end{center}
        \end{small}
\vspace{-2mm}
\end{table}

We can see that by using WER as the detection metric, the temporal dependency based method can achieve AUC as high as 0.936 on Common Voice and 0.93 on LIBRIS. We also explore different values of $k$ and we observe that the results do not vary too much (detailed results can be found in Table \ref{tab:parameter k} in Appendix).
When $k = 4 / 5$, the AUC score based on CER can reach $0.969$, which shows that such temporal dependency based method is indeed promising in terms of distinguishing adversarial instances. Interestingly, these results suggest that the temporal dependency based method would suggest an easy-implemented but effective method for characterizing adversarial audio attacks.

\begin{table}[t]
\caption{Evaluation of adaptive attacks }
\vspace{-2mm}
\label{tab:finaltable2}
\begin{center}
\begin{small}
\resizebox{0.85\textwidth}{23mm}{

\begin{tabular}{|c|c|c|c|c|}
\toprule
\multicolumn{3}{|c|}{Attack methods}                                                                                                                                & \multicolumn{2}{c|}{Optimization based attack (Opt)} \\ \hline
\multicolumn{3}{|c|}{Datasets}                                                                                                                                      & LibriSpeech            & CommonVoice           \\ \hline
\multicolumn{3}{|c|}{Evaluation metrics of adaptive attack}                                                                                                      & \multicolumn{2}{c|}{Attack success rate}                                                                \\ \hline
\multirow{3}{*}{\begin{tabular}[c]{@{}c@{}}Defense\\ Method\end{tabular}} & \multirow{3}{*}{\begin{tabular}[c]{@{}c@{}}Input\\ trans.\end{tabular}} & Down Samp. & 92\%                                                & 90\%                                                \\ \cline{3-5}
                                                                          &                                                                         & Quan-256   & 98\%                                                & 100\%                                               \\ \cline{3-5}
                                                                          &                                                                         & Median-4   & 98\%                                                & 96\%                                                \\ \hline
\multicolumn{3}{|c|}{Evaluation metrics for detection}                                                                                                           & \multicolumn{2}{c|}{AUC score}                                                                          \\ \hline
\multicolumn{3}{|c|}{Detection results of TD Method}                                                                                                             & 0.930                                              & 0.936                                              \\ \hline
\multicolumn{3}{|c|}{Segment attack}                                                                                                                             & 2\% success rate                                    & 2\% success rate                                    \\ \hline
\multicolumn{3}{|c|}{Concatenation attack}                                                                                                                       & Failed.                                            & Failed.                                            \\ \hline
\multicolumn{3}{|c|}{Combination attack under both random $k_A$ and $k_D$}                                                                                                                         & 0.873                                              & 0.877                                              \\ \hline
\end{tabular}}
\end{small}

\end{center}
\end{table}

\begin{table}[t]

\begin{small}
\begin{center}

\caption{AUC of detecting Combination Attack based on TD method}
\label{tab:comparison}
\begin{tabular}{ccccc}
% \toprule
\toprule
Combination & Detection  & \multicolumn{3}{c}{TD metrics}  \\ \cline{3-5}
                          Attack & Parameter $k_D$            & WER & CER & LCP            \\ \hline
\multirow{3}{*}{$k_A=\{\frac{1}{2}\}$} & 1/2                                       & 0.607 & 0.518 & 0.643 \\  \cline{2-5}
                          & 2/3                                      & 0.957 & 0.965 & 0.881 \\ \cline{2-5}
                          & Rand(0.2, 0.8)                                     & 0.889 & 0.882 & 0.776 \\ \hline

%\multirow{4}{*}{$k_A=\{\frac{1}{2}, \frac{2}{3} \}$}    & 1/2                                       & 0.633 & 0.690 & 0.552 \\ \cline{2-5}
%                          & 2/3                                      & 0.536 & 0.615 & 0.524 \\ \cline{2-5}
%                          & 3/4                                    & 0.942 & 0.974 & 0.934 \\ \cline{2-5}
%                          & Rand(0.2, 0.8)                                     & 0.801 & 0.880 & 0.664 \\ \hline

\multirow{4}{*}{$k_A=\{\frac{1}{2}, \frac{2}{3}, \frac{3}{4}\}$}    & 1/2                                       & 0.665	& 0.682 & 0.604 \\ \cline{2-5}
                          & 2/3                                      & 0.653	& 0.664 & 0.564 \\ \cline{2-5}
                          & 3/4                                    & 0.633 & 0.653 & 0.601
 \\ \cline{2-5}
                          & Rand(0.2, 0.8)                                     & 0.785 & 0.832 & 0.642 \\
%\multirow{4}{*}{$k_1=1/2$, $k_2=2/3$}    & 1/2                                       & 0.932 & 0.912 & 0.860 \\ \cline{2-5}
%                          & 2/3                                      & \textbf{0.611}	& \textbf{0.543} & \textbf{0.604} \\ \cline{2-5}
 %                         & 3/4                                    & 0.956	& 0.944 & 0.872 \\ \cline{2-5}
 %                         & Rand(0.2, 0.8)                                     & 0.879	& 0.891 & 0.698 \\
 \hline
\end{tabular}
\end{center}
        \end{small}
\end{table}

\subsection{Adaptive Attacks Against Defense and Detection methods}
In this section we measured some adaptive attack against the defense and detection methods.
Since the autoencoder based defense almost fails to defend against different attacks, here we will focus on the input transformation based defense and TD detection.
Given that Opt is the strongest attack here, we will mainly apply Opt to perform adaptive attack against the speech-to-text translation task. We list our experiments' structure in Table~\ref{tab:finaltable2}. For full results please refer to the Appendix.

\textbf{Adaptive Attacks Against Input Transformations}
\label{sec:adaptive}
Here we apply adaptive attacks against the preceding input transformations and therefore evaluate the robustness of the input transformation as defenses.
We implemented our adaptive attack based on three input transformation methods: Quantization, Local smoothing, and Downsampling.
For these transformation, we leverage a gradient-masking aware approach to
generate adaptive attacks.
% all the input transformation under on our adaptive attack method seemed to be totally ineffectiveness. As for autoencoder transformation method, since it does not work well, the adaptive attack is not meaningful enough.

In the optimization based attack~\citep{carlini2018audio}, the attack achieved by solving the optimization problem:
$\min_{\delta}\|\delta \|_{2}^{2} + c\cdot l(x+\delta, t)$,
where $\delta$ is referred as the perturbation, $x$ the benign audio, $t$ the target phrase, and $l(\cdot)$ the CTC-loss. Parameter $c$ is iterated to trade off the importance of being adversarial and remaining close to the original instance.

For quantization transformation, we assume the adversary knows the quantization parameter $q$. We then change our attack targeted optimization function to: $\min_{\delta} \|q\delta \|_{2}^{2} + c\cdot l(x+q\delta, t)$. After that, all the adversarial audios can be resistant against quantization transformations and it only increased a small magnitude of adversarial perturbation, which can be ignored by human ears. When $q$ is large enough, the distortion would increase but the transformation process is also ineffective due to too much information loss.

For downsampling transformation, the adaptive attack is conducted by performing the attack on the sampled elements of origin audio sequence. Since the whole process is differentiable, we can do adaptive attack through gradient directly and all the adversarial audios are able to attack.

For local smoothing transformation, it is also differentiable in case of average smoothing transformation, so we can pass the gradient effectively. To attack against median smoothing transformation, we can just convert the gradient back to the median and update its value, which is similar to the maxpooling layer's back propagation process. By implementing the adaptive attack, all the smoothing transformation is shown to be ineffective.

We chose our samples randomly from LIBRIS and Common Voice audio dataset with 50 audio samples each. We implemented our adaptive attack on the samples and passed them through the corresponding input transformation. We use down-sampling from 16kHZ to 8kHZ, median / average smoothing with one-sided sequence length $K = 4$, quantization method with $q = 256$ as our input transformation methods. In \citep{carlini2018audio}, Decibels (a logarithmic scale that measures the relative loudness of an audio sample) is applied as the measurement of magnitude of perturbation: $dB(x) = \max_{i}20\cdot \log_{10}(x_i)$, which $x$ referred as adversarial audio sampled sequence.
The relative perturbation is calculated as $dB_x(\delta) = dB(\delta) - dB(x)$, where $\delta$ is the crafted adversarial noise.
% \[dB(x) = \max_{i}20\cdot \log_{10}(x_i)\]
% and the comparative distortion was computed by:
% \[dB_x(\delta) = dB(\delta) - dB(x)\]

We measured our adaptive attack based on the same criterion. We show that all the adaptive attacks become effective with reasonable perturbation, as shown in Table~\ref{tab:evaluation of adaptive attack}.
As suggested in~\citep{carlini2018audio}, almost all the adversarial audios have distortion $dB_x(\delta)$ from -15dB to -45dB which is tolerable to human ears. From Table~\ref{tab:evaluation of adaptive attack}, the added perturbation are mostly within this range.
%We also listened to the adaptive attack samples and they are indeed hard to noticed by human auditory system.

\begin{table}[t]
        \caption{The $dB_x(\delta)$ evaluation of adaptive attack}
        \label{tab:evaluation of adaptive attack}
        \begin{center}
        \begin{small}
        \begin{tabular}{cccccc}
        \hline
        Dataset & Non-adaptive & Downsample & Quantization-256 & Median-4 & Average-4 \\
        \hline
        LIBRIS & -36.06 & -21.42 & -11.02 & -23.58 & -25.64\\
        CommmonVoice & -35.65 & -20.91 & -9.48 & -23.42 & -25.12\\
        \hline
        \end{tabular}
        \end{small}
        \end{center}
\end{table}

\textbf{Adaptive Attacks Against Temporal Dependency Based Method}
To thoroughly evaluate the robustness of temporal dependency based method, we also perform strong adaptive attack against it.
Notably, even if the adversary knows $k$, the adaptive attack is hard to conduct due to the fact that this process is non-differentiable. Therefore, we propose three types of strong adaptive attacks here aiming to explore the robustness of the temporal based method.
% In adaptive attack experiment of our detection method, making a adversarial audio which can recognized as our target by speech recognition network, but behaved normally after extract the first $k$-portion of it which is seemed to be difficult, even attacker can get out parameter $k$. The reason is that the detection itself is a procedure which has no gradient descending. We proposed two adaptive attack method:

    \emph{\textbf{Segment attack}}: Given the knowledge of $k$, we first split the audio into two parts: the prefix of length $k$ of the audio \sk~and the rest \slk.
    We then apply similar attack to add perturbation to only \sk. We hope this audio can be attacked successfully without changing \slk since the second part would not receive gradient updates. Therefore, when performing the temporal based consistency check, $T$(\sk) would be translated consistently with $T$(\wholek).

    % Then we implemented our audio attack normally, but only the first part $s1$ can obtain the gradient descending. We hope this audio can be attacked successfully without changing the rest part $s2$. So after the detection method implemented, the first part can remained unchanged due to the attack's target.
    \emph{\textbf{Concatenation attack}}: To maximally leverage the information of $k$, here we propose two ways to attack both \sk~and \slk~individually, and then concatenate them together.

            \textbf{1.} the target of \sk~ is the first $k-$portion of adversarial target, and \slk~ is attacked to the rest.

            \textbf{2.} the target of \sk~is the whole adversarial target, while we attack \slk~to be silence, which means \slk~transcribing nothing. This is different from segment attack where \slk~is not modified at all.

    \emph{\textbf{Combination attack}}: To balance attack success rate for both sections and the whole sentence against TD, we apply the attack objective function as $\min_{\delta}\|\delta \|_{2}^{2} + c\cdot (l(x+\delta, t) + l((x + \delta)_k, t_k)$, where $x$ refers to the whole sentence.
    % we hope this attack can balance importance of both attack target and against TD method without occuring too much distortion.

% The result of both adaptive attack is not promising. For Partial modification we found that in most consequences, attack itself is not successful by only do the modification on first part $s1$ and here's some attack sample in \ref{tab:partial}.
% Even in some special case that attack is successful, the first part still can't remain unchanged after separated from back part. We considered this case for some reasons: 1. First part is not robust enough to resist the temporal dependency even it has controlled the whole adversarial attack. 2. The speech recognition model based on biRNN will transfer the significance of $s2$ to disturb the whole $s1$ recognition procedure.

For segment attack, we found that in most cases the attack cannot succeed, that the attack success rate remains at $2\%$ for 50 samples in both LIBRIS and Common Voice datasets, and some of the examples are shown in Appendix.
We conjecture the reasons as: 1. \sk~alone is not enough to be attacked to the adversarial target due to the temporal dependency; 2. the speech recognition results on \slk~cannot be applied to the whole recognition process and therefore break the recognition process for \sk.

For concatenation attack, we also found that the attack itself fails. That is, the transcribed result of $adv$(\sk)+$adv$(\slk) differs from the translation result of \sk+\slk. Some examples are shown in Appendix.
% We considered this case as the temporal dependency will associate with not only the audio separated process, but the audio concat process. When two adversarial audio combined to a new audio, both audio will lost their adversarial ability partially revealing the ground truth sentence. Here's some samples in \ref{tab:separated}.
The failure of the concatenation adaptive attack more explicitly shows that the temporal dependency plays an important role in audio. Even if the separate parts are successfully attacked into the target, the concatenated instance will again totally break the perturbation and therefore render the adaptive attack inefficient. On the contrary, such concatenation will have negligible effects on benign audio instances, which provides a promising direction to detect adversarial audio.

For combination attack, we vary the section portion $k_D$ used by TD and evaluate the cases where the adaptive attacker uses the same/different section $k_A$. We define Rand(a,b) as uniformly sampling from [a,b].
% and $k_A$ refers to attacker's $k$ when $k_D$ refers to defender's.
We consider stronger attacker, for whom the $k_A$ can be a set containing random sections. The detection results for different settings are shown in Table~\ref{tab:comparison}.
From the results we can see that when $|k_A|=1$, if the attacker uses the same $k_A$ as $k_D$
to perform adaptive attack, the attack can achieve relative good performance and if attacker uses different $k_A$, the attack will fail with AUC above 85\%. We also evaluate the case that defender randomly sample $k_D$ during the detection and find that it's very hard for adaptive attacker to perform attacks, which can improve model robustness in practice. For $|k_A|>1$, the attacker can achieve some attack success when the set contains $k_D$. But when $|k_A|$ increases, the attacker's performance becomes worse. The complete results are given in Appendix. Notably, the random sample based TD appears to be robust in all cases.

%\begin{figure}[htbp]
%        \vspace{-2mm}
%        \includegraphics[height=2.8cm]{table1.png}
%        \caption{Examples for (a) segment attack and (b) concatenation attack.}
%label{tab:expattack}
%        \vspace{-4mm}
%\end{figure}

\section{Conclusion}
This papers proposes to exploit the temporal dependency property in audio data to characterize audio adversarial examples. Our experimental results show that while four primitive input transformations on audio fail to withstand adaptive adversarial attacks, temporal dependency is shown to be resistant to these attacks. We also demonstrate the power of temporal dependency for characterizing adversarial examples generated by
three state-of-the-art audio adversarial attacks. The proposed method is easy to operate and does not require model retraining. We believe our results shed new lights in exploiting unique data properties toward adversarial robustness.

%\bibliographystyle{iclr2019_conference}
%\bibliography{audio_defense}

%\newpage

% \section{Acknowledgement}
% We thank the feedback from Nicholas Carlini. This work is supported by
\newpage
\section*{Appendix}

\setcounter{equation}{0}
\setcounter{figure}{0}
\setcounter{table}{0}
\makeatletter
\renewcommand{\theequation}{A\arabic{equation}}
\renewcommand{\thefigure}{A\arabic{figure}}
\renewcommand{\thetable}{A\arabic{table}}

\subsection{Results on ``autoencoder transformation method for speech-to-text attack'' and ``Primitive transformation for speech-to-text attack''}

\begin{table}[htbp]
        \caption{Evaluation on Common Voice with language model}
        \label{tab:commondeflanguage}
        \begin{center}
        \begin{small}
        \begin{tabular}{ccccc}
        \hline
        Transformation Methods & OriginWER(\%) & OriginCER(\%) & AdvWER(\%) & AdvCER(\%)\\
        \hline
        Without transformations & 27.5 & 14.3 & 95.9 & 80.1 \\
        Autoencoder & 57.6 (2.09) & 34.1 (2.38) & 76.5 (0.80) & 49.8 (0.62)\\
        Median-4 & 27.0 (0.98) & 14.6 (1.02) & 73.6 (0.77) & 42.4 (0.53) \\
        Downsample & 31.2 (1.13) & 17.6 (1.23) & 69.6 (0.73) & 41.2 (0.51) \\
        Quan-128 & 34.4 (1.25) & 21.3 (1.49) & 75.9 (0.79) & 45.3 (0.57) \\
        Quan-256 & 42.9 (1.56) & 26.7 (1.87) & 70.7 (0.74) & 41.8 (0.52) \\
        Quan-512 & 52.4 (1.90) & 37.1 (2.59) & 68.5 (0.71) & 45.0 (0.56) \\
        Quan-1024 & 62.4 (2.27) & 47.2 (3.3) & 70 (0.73) & 51.2 (0.64) \\

        \hline
        \end{tabular}
        \end{small}
        \end{center}
        \vspace{-6mm}
\end{table}

\begin{table}[htbp]
        \caption{Evaluation on LIBRIS with language model}
        \label{tab:librideflanguage}
        \begin{center}
        \begin{small}
        \begin{tabular}{ccccc}
        \hline
        Transformation Methods & OriginWER(\%) & OriginCER(\%) & AdvWER(\%) & AdvCER(\%) \\
        \hline
        Without transformations & 3.05 & 1.46 & 102.8 & 86.5 \\
        Autoencoder & 30.0 (9.84) & 15.1 (10.34) & 99.4 (0.97) & 58.1 (0.67)\\
        Median-4 & 3.6 (1.18) & 1.7 (1.16) & 35.1 (0.34) & 19.0 (0.22) \\
        Downsample & 11.8 (3.87) & 5.7 (3.90) & 41.2 (0.40) & 21.8 (0.25) \\
        Quan-128 & 3.2 (1.04) & 1.5 (1.03) & 49.7 (0.48) & 28.2 (0.33) \\
        Quan-256 & \textbf{3.5 (1.13)} & \textbf{1.7 (1.16)} & \textbf{29.1 (0.28)} & \textbf{15.4 (0.18)} \\
        Quan-512 & 12.0 (3.93) & 6.6 (4.52) & 25.1 (0.24) & 13.3 (0.15) \\
        Quan-1024 & 30.7 (10.06) & 20.3 (13.90) & 36.6 (0.36) & 24.1 (0.28) \\

        \hline
        \end{tabular}
        \end{small}
        \end{center}
        %\vspace{-6mm}
        \vspace{-6mm}
\end{table}
\begin{table}[htbp]
        \caption{Evaluation on Common Voice without passing through language model}
        \label{tab:commonvoicedef}
        \begin{center}
        \begin{small}
        \begin{tabular}{ccccc}
        \hline
        Transformation Methods & OriginWER(\%) & OriginCER(\%) & AdvWER(\%) & AdvCER(\%) \\
        \hline
        Without transformations & 37.7 & 18.5 & 95.8 & 83.0 \\
        Median-4 & 43.4 (1.15) & 20.4 (1.10) & 83.0 (0.87) & 46.5 (0.56) \\
        Down sampling & 47.2 (1.25) & 23.3 (1.26) & 77.6 (0.81) & 43.9 (0.53) \\
        Quantization-128 & 47.3 (1.25) & 25.7 (1.39) & 80.7 (0.84) & 49.0 (0.59) \\
        Quantization-256 & 52.5 (1.39) & 29.2 (1.58) & 73.4 (0.77) & 43.6 (0.53) \\
        Quantization-512 & 64.1 (1.70) & 37.5 (2.03) & 73.7 (0.77) & 44.2 (0.53) \\
        Quantization-1024 & 72.1 (1.91) & 50.4 (2.72) & 76.9 (0.80) & 53.0 (0.64) \\
        \hline
        \end{tabular}
        \end{small}
        \end{center}
        \vspace{-6mm}
\end{table}

\begin{table}[!htbp]
        \caption{Evaluation on LIBRIS without passing through language model}
        \label{tab:librispeechdef}
        \begin{center}
        \begin{small}
        \begin{tabular}{ccccc}
        \hline
        Transformation Methods & OriginWER(\%) & OriginCER(\%) & AdvWER(\%) & AdvCER(\%) \\
        \hline
        Without transformations & 12.4 & 7.05 & 105.3 & 91.7 \\
        Median-4 & 16.4 (1.32) & 8.0 (1.13) & 57.9 (0.55) & 27.5 (0.30) \\
        Downsample & 24.2 (1.95) & 13.0 (1.84) & 60.9 (0.58) & 31.2 (0.34) \\
        Quantization-128 & 13.4 (1.08) & 7.6 (1.08) & 66.1 (0.63) & 37.1 (0.40) \\
        Quantization-256 & \textbf{16.3 (1.31)} & \textbf{8.9 (1.26)} & \textbf{48.6 (0.46)} & \textbf{24.0 (0.26)} \\
        Quantization-512 & 27.5 (2.21) & 13.8 (1.96) & 47.0 (0.45) & 23.0 (0.25) \\
        Quantization-1024 & 46.8 (3.77) & 25.4 (3.60) & 52.3 (0.50) & 30.0 (0.33) \\
        \hline
        \end{tabular}
        \end{small}
        \end{center}
\end{table}

\newpage
\subsection{More results on primitive transformation method for audio classification attack}

\begin{figure}[htbp]
    \begin{minipage}[h]{0.55\textwidth}
        \includegraphics[height=7cm]{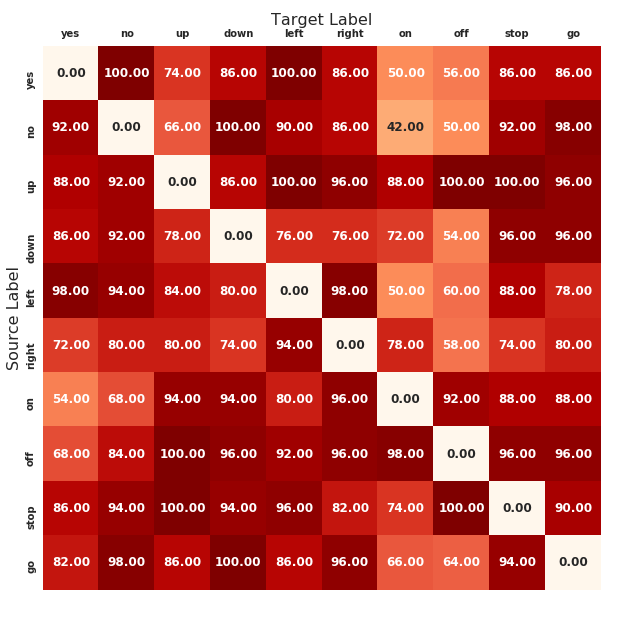}
        \caption{Successful attack rates}
        \label{tab:Pic1}
    \end{minipage}
    \begin{minipage}[h]{0.55\textwidth}
        \includegraphics[height=7cm]{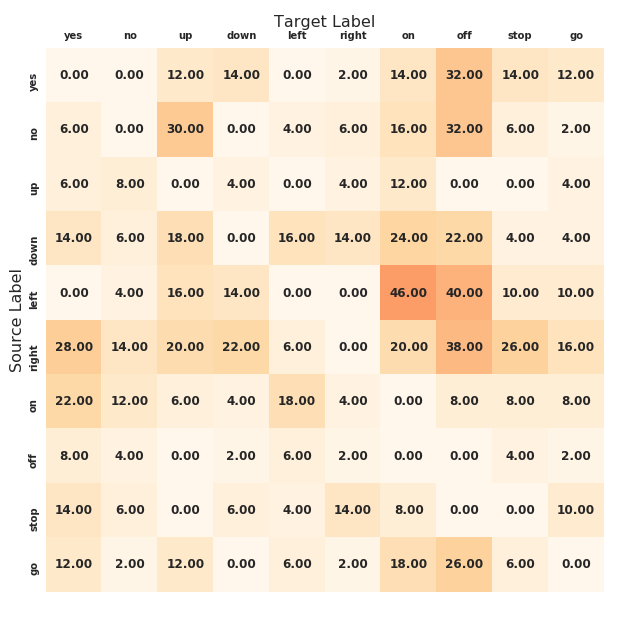}
        \caption{Unchanged label rates}
        \label{tab:Pic2}
    \end{minipage}
\end{figure}

\begin{figure}[htbp]
    \begin{minipage}[h]{0.55\textwidth}
        \includegraphics[height=7cm]{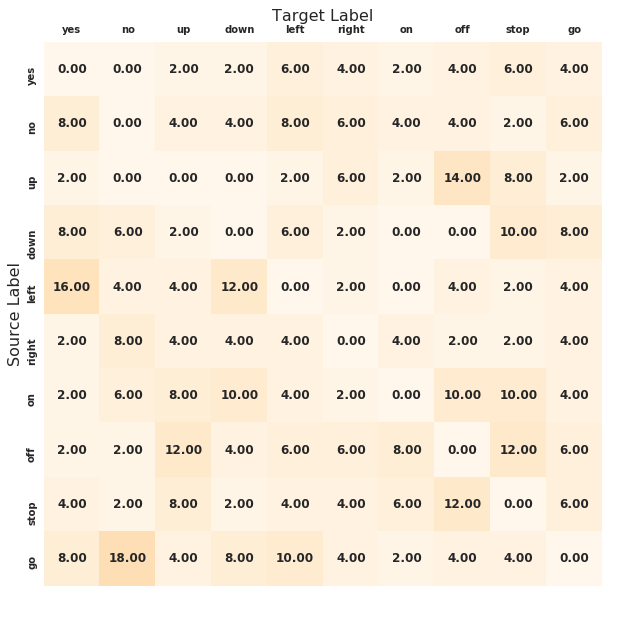}
        \caption{Successful attack rates after transformation}
        \label{tab:Pic3}
    \end{minipage}
    \begin{minipage}[h]{0.55\textwidth}
        \includegraphics[height=7cm]{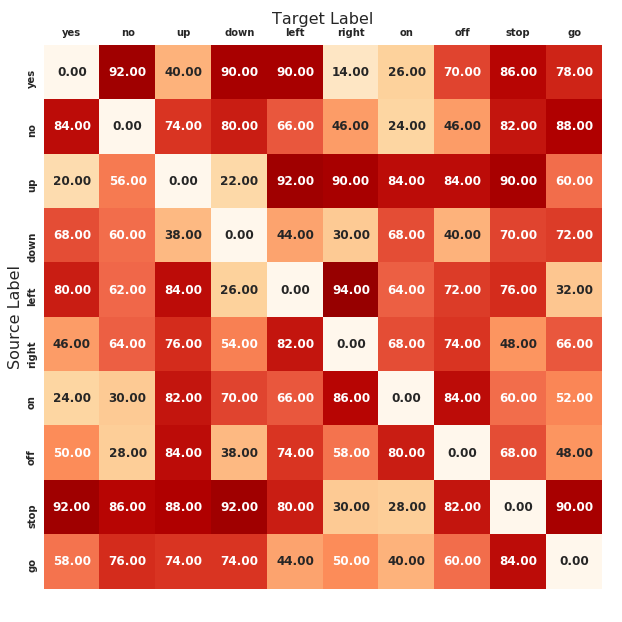}
        \caption{Unchanged label rates after transformation}
        \label{tab:Pic4}
    \end{minipage}
\end{figure}

\subsection{More results on adaptive attacks against temporal dependency based method}

\begin{table}[htbp]
\begin{small}
\vspace{-8mm}
        \caption{Examples of Segment Attack and Concatenation attack}
        \label{tab:partial}
        \begin{center}
        \begin{small}
        \begin{tabular}{ll}
        \hline
        Type & Transcribed results \\
        \hline
        Original & and he leaned against the wa lost in reveriey \\
        the first half of Original & and he leaned against the wa\\ \\

        Adaptive attack target & this is an adversarial example\\
        Adaptive attack result & this is an adversarial {\color{red}losin ver}\\
        the first half of Adv. & this is a {\color{red}agamsa}\\ \\

        Adaptive attack target & okay google please cancel my medical appointment\\
        Adaptive attack result & okay google please cancel my {\color{red}medcalosinver}\\
        the first half of Adv. & okay {\color{red}go} please\\

        \hline\hline
        Original & why one morning there came a quantity of people and set to work in the loft \\
        Attack target & this is an adversarial example\\ \\

        \sk & this is an\\
        \slk & adversarial example\\
        \sk+\slk & this is {\color{red}a quantity of people and set to work in a lift}\\ \\

        \sk & this is an adversarial example\\
        \slk & $sil$\\
        \sk+\slk & this is an {\color{red}adernari eanquatete of pepl and sat to work in the loft}\\
        \hline
        \end{tabular}
        \end{small}
        \end{center}

\vspace{-2mm}
        \end{small}
\end{table}
%\begin{figure}[htbp]
%        \vspace{-2mm}
%        \includegraphics[height=2.8cm]{table1.png}
%        \caption{Examples for (a) segment attack and (b) concatenation attack.}
%label{tab:expattack}
%        \vspace{-4mm}
%\end{figure}
\begin{table}[htb]
\begin{center}
\begin{small}
        \vspace{-6mm}
        \caption{AUC scores of different $k$}
        \label{tab:parameter k}
        \begin{center}
        \begin{tabular}{cccccc}
        \hline
        $k$ & WER & CER & LCP\\
        \hline
        $1/2$ & 0.930 & 0.933 & 0.806\\
        $2/3$ & 0.930 & 0.948 & 0.826\\
        $3/4$ & 0.933 & 0.938 & 0.839\\
        $4/5$ & 0.955 & \textbf{0.969} & 0.880 \\
        $5/6$ & 0.941 & 0.962 & 0.858 \\
        \hline
        \end{tabular}
        \end{center}

        \vspace{-6mm}
\end{small}
\end{center}
\end{table}

\begin{table}[htb]
\begin{center}
\caption{AUC of detecting Combination Attack based on TD method}
\label{tab:comparison2}
\begin{tabular}{ccccc}
% \toprule
\toprule
Combination & Detection  & \multicolumn{3}{c}{TD metrics}  \\ \cline{3-5}
                          Attack & Parameter $k_D$            & WER & CER & LCP            \\ \hline
\multirow{4}{*}{$k_A=\{\frac{1}{2}\}$} & 1/2                                       & 0.607 & 0.518 & 0.643 \\  \cline{2-5}
                          & 2/3                                      & 0.957 & 0.965 & 0.881 \\ \cline{2-5}
                          & 3/4                                     & 0.943	& 0.951 & 0.875 \\  \cline{2-5}
                          & Rand(0.2, 0.8)                                     & 0.889 & 0.882 & 0.776 \\ \hline
\multirow{4}{*}{$k_A=\{\frac{2}{3}\}$}    & 1/2                                       & 0.932 & 0.912 & 0.860 \\ \cline{2-5}
                          & 2/3                                      & 0.611	& 0.543 & 0.604 \\ \cline{2-5}
                          & 3/4                                    & 0.956	& 0.944 & 0.872 \\ \cline{2-5}
                          & Rand(0.2, 0.8)                                     & 0.879	& 0.890 & 0.762 \\ \hline
\multirow{4}{*}{$k_A=\{\frac{1}{2}, \frac{2}{3} \}$}    & 1/2                                       & 0.633 & 0.690 & 0.552 \\ \cline{2-5}
                          & 2/3                                      & 0.536 & 0.615 & 0.524 \\ \cline{2-5}
                          & 3/4                                    & 0.942 & 0.974 & 0.934 \\ \cline{2-5}
                          & Rand(0.2, 0.8)                                     & 0.801 & 0.880 & 0.664 \\ \hline
\multirow{4}{*}{$k_A=\{\frac{1}{2}, \frac{2}{3}, \frac{3}{4}\}$}    & 1/2                                       & 0.665	& 0.682 & 0.604 \\ \cline{2-5}
                          & 2/3                                      & 0.653	& 0.664 & 0.564 \\ \cline{2-5}
                          & 3/4                                    & 0.633 & 0.653 & 0.601
 \\ \cline{2-5}
                          & Rand(0.2, 0.8)                                     & 0.785 & 0.832 & 0.642 \\ \hline \multirow{4}{*}{$k_A=\{\frac{1}{2}, \frac{2}{3}, \frac{3}{4}, \frac{4}{5}\}$}    & 1/2                                       & 0.701	& 0.712 & 0.615 \\ \cline{2-5}
                          & 2/3                                      & 0.684	& 0.701 & 0.583 \\ \cline{2-5}
                          & 3/4                                    & 0.681 & 0.693 & 0.613
 \\ \cline{2-5}
                          & Rand(0.2, 0.8)                                     & 0.742 & 0.811 & 0.623 \\ \hline
\multirow{4}{*}{$k_A=\{\frac{1}{2}, \frac{2}{3}, \frac{3}{4}, \frac{4}{5}, \frac{5}{6} \}$}    & 1/2                                       & 0.736	& 0.784 & 0.601 \\ \cline{2-5}
                          & 2/3                                      & 0.723	& 0.763 & 0.612 \\ \cline{2-5}
                          & 3/4                                    & 0.715 & 0.755 & 0.584
 \\ \cline{2-5}
                          & Rand(0.2, 0.8)                                     & 0.734 & 0.801 & 0.620 \\ \hline
\multirow{4}{*}{$k_A= \text{Rand(0.2, 0.8)}$}    & 1/2                                       & 0.880	& 0.881 & 0.824 \\ \cline{2-5}
                          & 2/3                                      & 0.922	& 0.972 & 0.831 \\ \cline{2-5}
                          & 3/4                                    & 0.952 & 0.968 & 0.894
 \\ \cline{2-5}
                          & Rand(0.2, 0.8)                                     & 0.873 & 0.875 & 0.799 \\ \hline
%\multirow{4}{*}{$k_1=1/2$, $k_2=2/3$}    & 1/2                                       & 0.932 & 0.912 & 0.860 \\ \cline{2-5}
%                          & 2/3                                      & \textbf{0.611}	& \textbf{0.543} & \textbf{0.604} \\ \cline{2-5}
 %                         & 3/4                                    & 0.956	& 0.944 & 0.872 \\ \cline{2-5}
 %                         & Rand(0.2, 0.8)                                     & 0.879	& 0.891 & 0.698 \\ \hline
\end{tabular}
\end{center}

\end{table}

% \usepackage{multirow}
% \usepackage{multirow}

\end{document}